\documentclass[10pt,twocolumn]{article} 
\usepackage{times}
\usepackage{graphicx}
\usepackage{amssymb}
\usepackage{url,hyperref}
\usepackage{color,soul}
\usepackage{adjustbox}
\usepackage{amsmath}
\makeatletter
\newcommand{\printfnsymbol}[1]{%
  \textsuperscript{\@fnsymbol{#1}}%
}
\makeatother

\date{}

\begin{document}

\title{\vspace{-1.5em}Producing augmentation-invariant embeddings \\
	 from real-life imagery\vspace{1em}}
% Supporting large-scale instance recognition with out-of-domain images
\author{
Sergio Manuel Papadakis \\\
  \texttt{socom2.00@gmail.com} \\
\and
Sanjay Addicam \\
  \texttt{addicam.sanjay@gmail.com} \\
}

\maketitle

\begin{abstract}
This article presents an efficient way to produce feature-rich, high-dimensionality embedding spaces from real-life images. 
The features produced are designed to be independent from augmentations used in real-life cases which appear on social media.
Our approach uses convolutional neural networks (CNN) to produce an embedding space.
An ArcFace head \cite{arcface2019v3} was used to train the model by employing automatically produced augmentations. 
Additionally, we present a way to make an ensemble out of different embeddings containing the same semantic information, a way to normalize the resulting embedding using an external dataset, and a novel way to perform quick training of these models with a high number of classes in the ArcFace head.
Using this approach we achieved the 2nd place in the 2021 Facebook AI Image Similarity Challenge: Descriptor Track  \cite{competition2021url}.
\end{abstract}

\section{Introduction}

The Facebook Image Similarity Dataset and Challenge is a benchmark for large-scale image similarity detection \cite{competition2021paper}. 
The dataset consists of $1$ million reference images and $100$ thousand query images ($50$ thousand for Phase 1 and $50$ thousand for Phase 2 of the competition).
In addition, it is provided a training set of $1$ million images through which it is possible to perform model training and score normalization.
Some of the query images are augmented versions of the reference images.
The competition's goal is to determine which of the query images were built out of the reference dataset.
Some of the challenges of this dataset are: the huge imbalance of classes, the variety and severity of the augmentations used in the query set, the prohibition in the use of the reference set for training purpose and the fact that the predicted similarity score of a given pair of images reference-query must be independent from other images in the reference and query datasets.

The description track of this competition asks the participants to predict a set of description embeddings for each of the reference and query images.
The embeddings must have a maximum of $256$ dimensions.
The similarity score for a given reference-query pair is calculated as the negative of the squared Euclidean distance between the embeddings of each pair of images.
To measure the overall performance of the model, the competition evaluates the micro Average Precision ($\mu AP$) calculated from the similarity scores \cite{competition2021paper}.

The competition was split into two phases.
In Phase 1, ground truth data was provided for the first $25$ thousand query samples and the other $25$ thousand query samples could be validated by using an online leader board. 
In Phase 2 a new set of $50$ thousand query samples was provided to carry out the final testing of the model.

This paper summarizes our 2nd place winning solution for this chanllenge. We also made the code publicly available  \footnote{\fontsize{7}{9}\selectfont{https://github.com/socom20/facebook-image-similarity-challenge-2021}}.

\section{Methodology}

\begin{figure}[h!]
	\begin{center}
		\includegraphics[width=1.0\columnwidth]{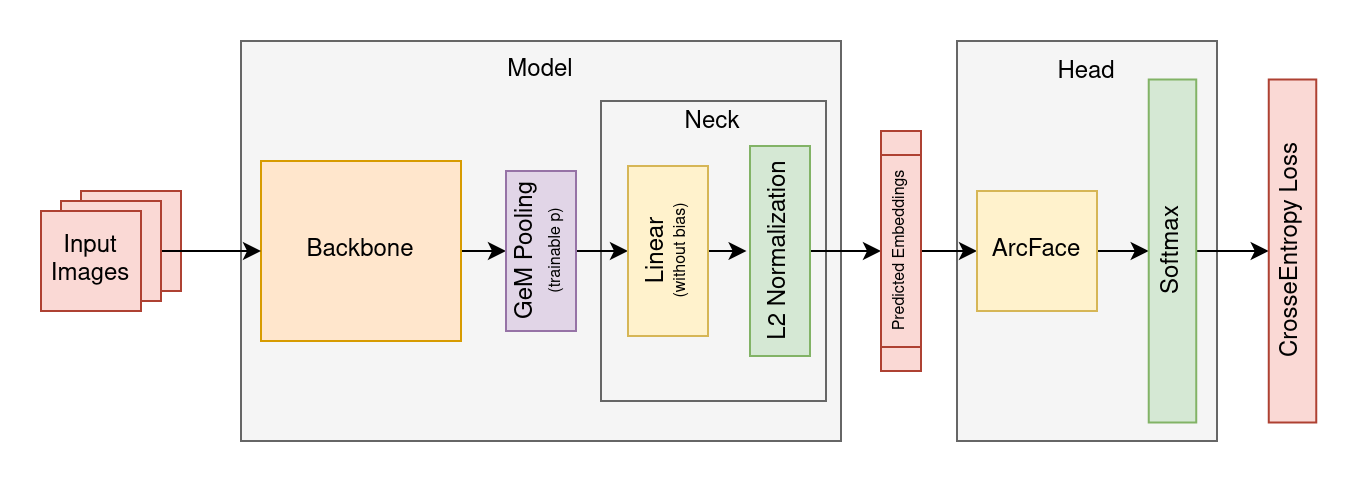}
	\end{center}
	\vspace*{-5mm}
	\caption{\small Model architecture, predicted embeddings and training loss function.}
	\label{model}
\end{figure}

\subsection{Global approach}
\label{sec:global_approach}

At a high level approach, the main problem of this challenge was the huge class imbalance ($50$k query images vs. $1$M reference images). It was immediately evident for us that we needed to reproduce the augmentations used in the query dataset to increase the training samples. For this purpose, we used Augly \cite{augly} and created some new augmentations to build synthetic query images.

In order to train computer vision models, we first tried using a triplet-based loss \cite{triplet}. However, searching for negative samples of sufficient quality became computationally difficult and the models' performance was not very good. Due to this problem we decided to use an ArcFace head to learn the image embeddings.

We train different models using different image sizes, different embedding dimensions, and different number of classes in the ArcFace head. These models were finally ensembled and the final embedding space was normalized using the competition's training dataset.

To validate the models during training we made use of the $25$ thousand ground truth samples provided in the competition's dataset.

\subsection{Modeling}
\label{sec:modeling}

We train different CNN backbones, always keeping the basic model architecture.
Figure \ref{model} shows the model architecture.
We used augmented versions of the training images as input to the CNN backbone. 
The produced features were pooled by using a Generalized Mean pooling layer (GeM), setting its \emph{p} parameter as trainable.
The pooled features were resized to the desired embedding dimension using a Linear layer without biases (we had observed that the bias term causes optimization problems during training).
The features are finally L2-normalized to produce the embedding representation of the input image.

In order to train the model we included an ArcFace head which produces softmax outputs.
Each class of the ArcFace head corresponds to a different augmented training image. We also vary the number of classes in each model. Table \ref{all_models} shows all the different models that were trained, always starting from a pre-trained backbone. The arc margin was fixed at $0.4$ and the softmax scaling factor to $40$.

\begin{table}[h]
	%   \centering
	\centering
	\begin{adjustbox}{width=\columnwidth,center}
		\begin{tabular}{|c|c|c|c|c|}
			\hline
			Backbone type & image size & Nº of classes & Embed dim. & Val. $\mu AP$ \\
			\hline
			EfficientNetV2 l &  512x512 &   980000 & 512 & 0.636 \\
			EfficientNetV2 s &  224x224 &  1200000 & 384 & 0.638 \\
			EfficientNetV2 s &  160x160 &  1000000 & 384 & 0.652 \\
			NfNet l1         &  160x160 &  1000000 & 384 & 0.664 \\
			NfNet l1         &  160x160 &  1500000 & 256 & 0.661 \\
			EfficientNet b5  &  160x160 &  1500000 & 256 & 0.652 \\
			EfficientNet b5  &  160x160 &  1800000 & 256 & 0.664 \\
			\hline
		\end{tabular}
	\end{adjustbox}
	\caption{All participating models in the final ensemble.}
	\label{all_models}
	
\end{table}

\subsection{External Datasets}
\label{sec:external_datasets}

In addition to the Facebook dataset we use two external datasets.

In order to increase the number of images utilized the ArcFace head, we used the ImageNet dataset \cite{imagenet}, which supplied us with $1.43$ million extra samples.

When performing partial validations of the models, we observed that they had trouble producing embeddings for images on people's faces. So as to try to fix this issue we decided to use the Deepfake Detection dataset \cite{deepfake}. We extracted $3$ frames from each video: an initial one, a middle one, and a last frame at the end of the videos.
This dataset provided us with $476$ thousand extra samples containing faces. Unfortunately this dataset provided us with many similar images since different videos were recorded from the same perspective. Due to the addition of these samples, the models' training became harder, so we did not finish utilizing many samples from this dataset and it did not help increasing the validation $\mu AP$ of the models. After experimenting with this dataset, we abandoned it and we did not use it in any of the final models.

\subsection{Augmentations}
\label{sec:augmentations}

\begin{figure}[t!]
	\begin{center}
		\includegraphics[width=1.0\columnwidth]{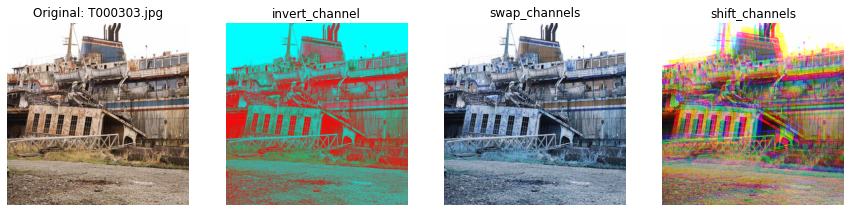}
	\end{center}
	\vspace*{-5mm}
	\caption{\small Example of the new augmentations created: \emph{invert\_channel}, \emph{swap\_channels} and \emph{shift\_channels}.}
	\label{new_augs}
\end{figure}

\begin{figure}[t!]
	\begin{center}
		\includegraphics[width=1.0\columnwidth]{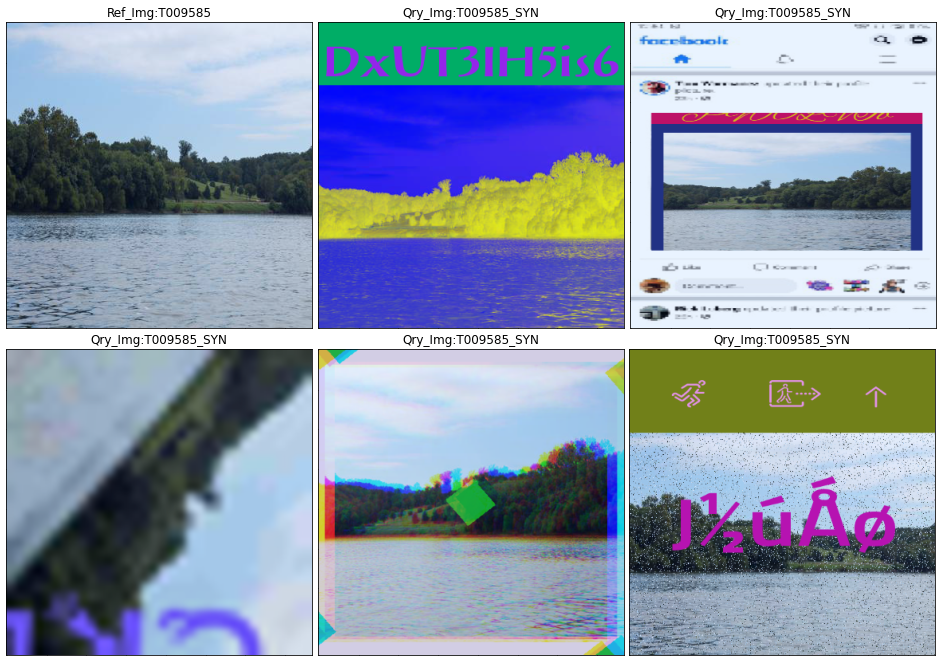}
	\end{center}
	\vspace*{-5mm}
	\caption{\small Examples of synthetically created query samples. The top right sample is the training image used to produce the augmentations.}
	\label{aug_samples}
\end{figure}

The augmentation step for this competition was crucial to improve our models' performance. 
We used all the augmentations available in the Augly library \cite{augly}.
Table \ref{all_augs} shows a list of all the augmentations used to build synthetic query images from the training samples.
From the inspection of query images, we observed 3 types of easy-to-implement augmentations not available in the Augly library, so we implemented them. 
An example of these new augmentations is shown in Figure \ref{new_augs}.
Finally, Figure \ref{aug_samples} shows examples of synthetically-created query samples.

\begin{table}[h]
	%   \centering
	\centering
	\begin{adjustbox}{width=0.8\columnwidth,center}
		\begin{tabular}{|ll|}
			\hline
			Augmentation name &  \\
			\hline
			scale                     &  overlay\_image   \\
			rotate                    &  blur             \\
			perspective\_transform    &  sharpen          \\
			encoding\_quality         &  shuffle\_pixels  \\
			color\_jitter             &  overlay\_stripes \\
			grayscale                 &  overlay\_emoji   \\
			opacity                   &  overlay\_text    \\
			crop                      &  vflip            \\
			pad                       &  hflip            \\
			pad\_square               &                   \\
			meme\_format              &  invert\_channel  \\
			overlay\_onto\_screenshot &  swap\_channels   \\
			pixelization              &  shift\_channels  \\
			\hline
		\end{tabular}
	\end{adjustbox}
	\caption{Different types of augmentations used.}
	\label{all_augs}
	
\end{table}

\subsection{Training method}
\label{sec:training_method}
To train the models we used \emph{Adam} optimizer with default parameters and categorical cross entropy as the loss function of the ArcFace output.
Due to the large memory requirement that ArcFace demands, we also tested \emph{SGD} optimizer which uses less memory than \emph{Adam}.
However, the training turned out to be much slower, so we decided to continue using Adam.
As learning rate schedulers we used \verb|ReduceLROnPlateau| and \verb|LinearWarmupCosineAnnealingLR|.

We also used different tricks to increase the training speed:

\begin{itemize}
\item As the training progressed, we increased the augmentation intensity.
This approach was adopted from \cite{effnetv2}.

\item We gradually increased input resolution as the training progressed \cite{effnetv2}.

\item Drip Training procedure: The idea was to iteratively increase the number of classes that the model is trained with. At the beginning of each iteration, the backbone is used to build a new centroids matrix of the ArcFace head. In section \ref{drip_training} we expand this method.

\end{itemize}

\begin{figure*}[h!]
	\begin{center}
		\includegraphics[width=5.5in]{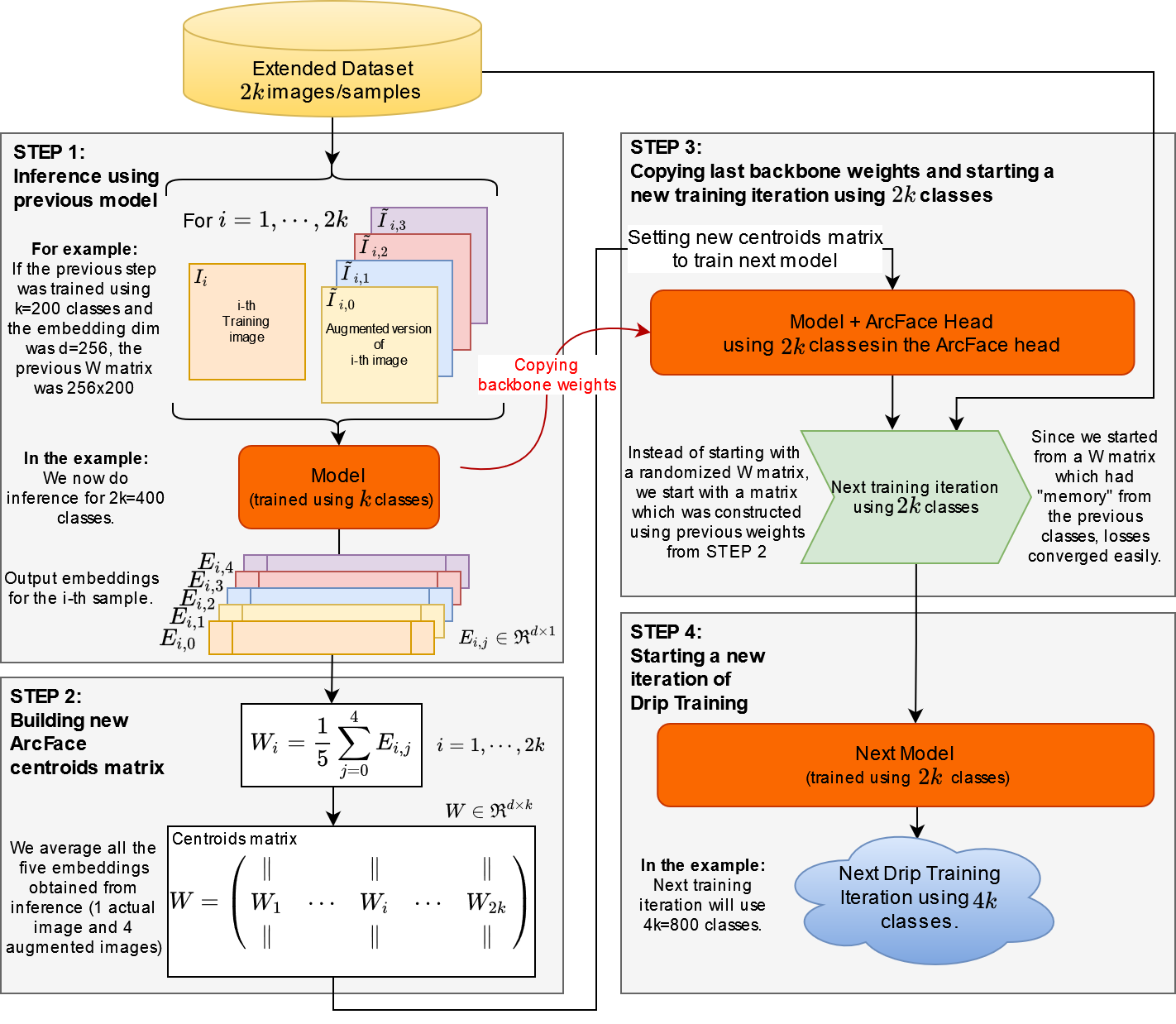}
	\end{center}
	\vspace*{-5mm}
	\caption{\small Examples of drip a training iteration.}
	\label{drip_training_fig}
\end{figure*}

\subsubsection{Drip Training}
\label{drip_training}

When we tried to approach the training using more than 1 million classes in the ArcFace head, faced with several convergence problems along the training. We found a way to gradually increase the complexity of the model during training, without forgetting the learned embeddings. Below we
display this training procedure:

\begin{enumerate}
\item We start training the model using a low number of classes, for instance $40$k target classes. By using this number of classes, a model can converge in an hour or so.

\item \label{item_start} After having reached a low-loss value ($loss < 1.0$), we stepped the number of output classes in the ArcFace head up to approximately twice its previous value.

\item To prevent catastrophic forgetting of the model, we used the backbone to predict embeddings of the new extended dataset. For each ArcFace class, we predicted $1$ embedding using the training image without augmentations, and $4$ embeddings augmenting the same sample.
These $5$ embeddings were finally averaged to produce one entry of the $W$ matrix of the ArcFace head (being $W$ the centroids matrix of ArcFace, shaped: $\verb|embedding_dim| \times \verb|class_number|$ \cite{arcface2019v3}).
We continued the embeddings calculation until we populated the entire $W$ matrix.

\item \label{item_end} Using the updated $W$ matrix, we started a new training. Since in this case we start from a low loss value the training do not take long to converge.

\item We repeated steps \ref{item_start} to \ref{item_end} until we completed a training using the desired number of
classes in the ArcFace head.

Figure \ref{drip_training_fig} shows a flux diagram of an iteration of Drip Training.

\end{enumerate}

\subsection{Ensembling}
\label{sec:ensembling}

\begin{figure}[t!]
	\begin{center}
		\includegraphics[width=1.0\columnwidth]{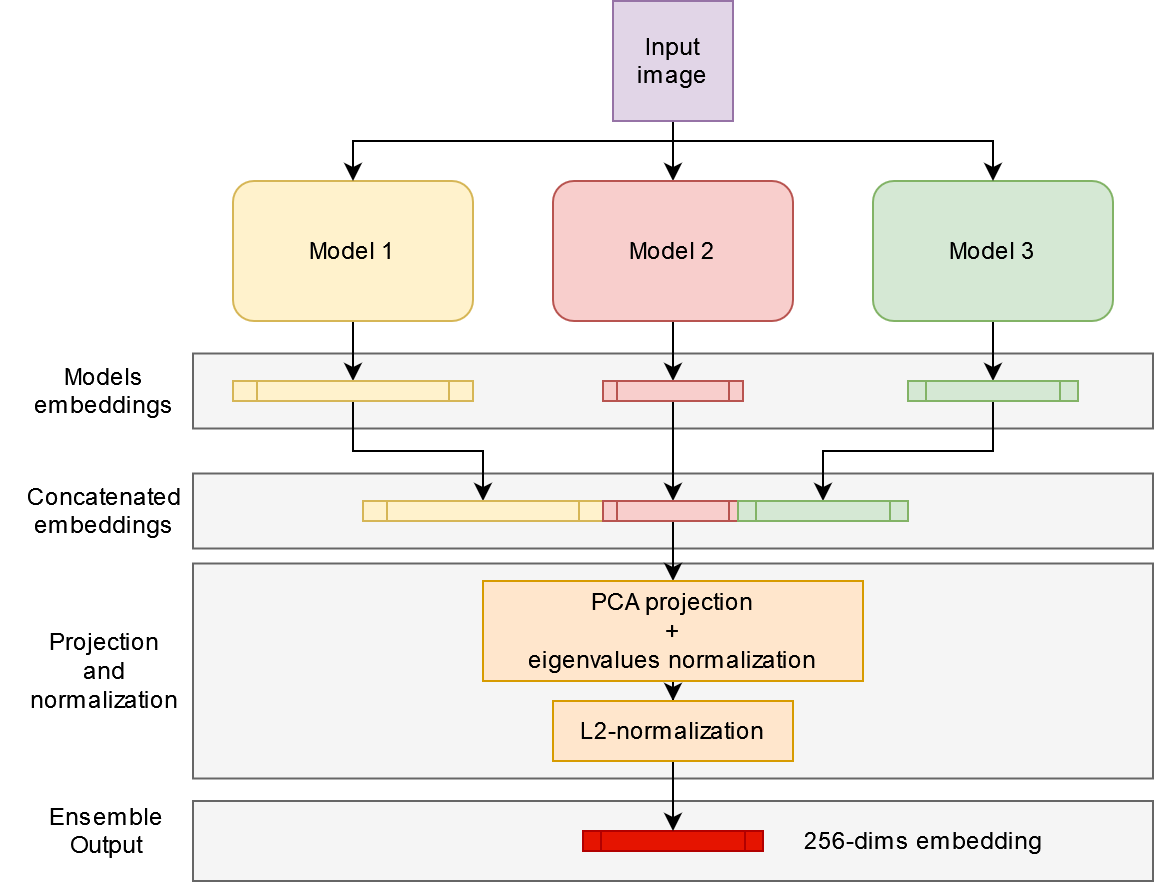}
	\end{center}
	\vspace*{-5mm}
	\caption{\small Examples of a three models ensemble.}
	\label{ensemble}
\end{figure}

As different models learn different embeddings for the same semantic information, we needed to project the different embeddings into a common vector-base among all the models before making the ensemble.
We used \emph{PCA} to project the embeddings.
Before projecting, we concatenated all models' embeddings in order to make one big embedding.
We used the un-augmented version of the training samples to train the \emph{PCA} algorithm making use of the Faiss library \cite{faiss}.
Then we selected the first 256 most important dimensions of the projected space to build the output embedding.

We tried several post-projection normalizations. We observed that: normalizing the output embedding space by the square root of the eigenvalues and a final L2-normalization of it was the best scoring approach.
Figure \ref{ensemble} shows an example of an ensemble of three models.

\subsection{Query embedding normalization}
\label{sec:normalization}

In order to produce an embedding normalization, we tried different strategies.
As the metric used by the competition is $\mu AP$, the similarity results for all query-reference pairs of images must be comparable.
One way to accomplish this objective is through a normalization step by using an external dataset. In our case we could use the training dataset.
As indicated by \cite{competition2021paper}, one way to achieve this normalization is as follows:

\begin{equation}
\label{eq_norm1}
\hat{S}(\vec{E}q_i,\vec{E}r_j) = S(\vec{E}q_i,\vec{E}r_j) -  \frac{\alpha}{n} \sum_{k=1}^{n} S( \vec{E}q_i, \vec{E}t_{i,k} )
\end{equation}

Where:
\begin{itemize}
\item $\vec{E}q_i$: is the embedding of the \emph{i-th} query image.
\item $\vec{E}t_{i,k}$: is the embedding of the \emph{k-th} most similar training image to $\vec{E}q_i$.
\item $\vec{E}r_j$: is the embedding of the \emph{j-th} reference image.
\item $S(\vec{E}_i,\vec{E}_j)$: is the similarity score between $\vec{E}_i$ and $\vec{E}_j$.
\item $\hat{S}(\vec{E}q_i,\vec{E}r_j)$: is the normalized similarity score between $\vec{E}q_i$ and $\vec{E}r_j$.
\item $\alpha$: is a constant factor.
\item $n$: is the number of closest training embeddings $\vec{E}t_{i,k}$ used to normalize $S(\vec{E}q_i,\vec{E}r_j)$.
\end{itemize}

As expression \ref{eq_norm1} normalizes similarity scores, it can not be used as such in the description track of the competition since it asks the participants to submit embeddings. 

Like it was mentioned before, the competition uses the negative of the squared Euclidean distance between the provided embeddings as the similarity score $S(\vec{E}_i,\vec{E}_j)$. Competition's similarity score can expressed as:

\begin{equation}
	\label{eq_sim}
	S(\vec{E}_i,\vec{E}_j) = - D^2(\vec{E}_i,\vec{E}_j) = - \| \vec{E}_i - \vec{E}_j \|^2_2
\end{equation}

Where:
\begin{itemize}
\item $D^2(\vec{E}_i,\vec{E}_j)$: is the squared Euclidean distance between $\vec{E}_i$ and $\vec{E}_j$.
\item $\| \cdot \|_2$: denotes the L2 norm function.
\end{itemize}

Replacing \ref{eq_sim} in \ref{eq_norm1}, the normalized distance between $\vec{E}q_i$ and $\vec{E}r_j$ can be expressed as:

\begin{eqnarray}
	\label{eq_norm2}
	\hat{D}^2(\vec{E}q_i,\vec{E}r_j) =   \| \vec{E}q_i - \vec{E}r_j \|^2_2 -  \frac{\alpha}{n} \sum_{k=1}^{n} \| \vec{E}q_i - \vec{E}t_{i,k} \|^2_2 \\
	\hat{D}^2(\vec{E}q_i,\vec{E}r_j) =  - \hat{S}(\vec{E}q_i,\vec{E}r_j)  
\end{eqnarray}

As the embeddings produced by our models and ensemble are L2-normalized, the following relation holds \cite{norm_face}:

\begin{equation}
	\label{eq_relation_euc_cos}
	 \| \vec{E}_i - \vec{E}_j \|^2_2 = 2\left[ 1 - C(\vec{E}_i, \vec{E}_j) \right]
\end{equation}

Where:
\begin{itemize}
\item $C(\vec{E}_i, \vec{E}_j)$: is the cosine similarity between $\vec{E}_i$ and $\vec{E}_j$.
\end{itemize}

Replacing \ref{eq_relation_euc_cos} in \ref{eq_norm2}, we get:

\begin{equation}
	\label{eq_norm4}
	\begin{split}
	\hat{D}^2(\vec{E}q_i,\vec{E}r_j) =  \| \vec{E}q_i - \vec{E}r_j \|^2_2 + \\ 
	+ 2 \frac{\alpha}{n} \sum_{k=1}^{n} C(\vec{E}q_i, \vec{E}t_{i,k} )  - 2 \alpha
	\end{split}
\end{equation}

By discarding the constant term $- 2 \alpha$ in \ref{eq_norm4} without losing the normalization power of the expression and arranging it, we finally get:

\begin{eqnarray}
	\label{eq_norm5}
		\hat{D'}^2(\vec{E}q_i,\vec{E}r_j) =  \| \vec{E}q_i - \vec{E}r_j \|^2_2 + \left[ \beta \sqrt{ \hat{C}(\vec{E}q_i) } \right] ^2   \\
		\label{eq_norm6}
		\hat{C}(\vec{E}q_i) = \frac{1}{n} \sum_{k=1}^{n} C(\vec{E}q_i, \vec{E}t_{i,k} )
\end{eqnarray}

\begin{figure}[t!]
	\begin{center}
		\includegraphics[width=0.8\columnwidth]{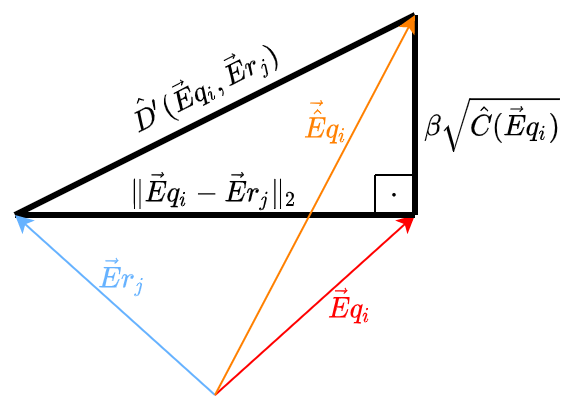}
	\end{center}
	\vspace*{-5mm}
	\caption{\small Triangular relationship defined by the normalization of the distance $\| \vec{E}q_i - \vec{E}r_j \|_2$.}
	\label{fig_trienagle}
\end{figure}

Where:
\begin{itemize}
	\item $\beta$: is a constant factor.
	\item $\hat{D'}^2(\vec{E}q_i,\vec{E}r_j)$: is the normalized squared Euclidean distance between $\vec{E}_i$ and $\vec{E}_j$.
\end{itemize}
	
From equation \ref{eq_norm5} we can note that it defines the triangular relationship shown in Figure \ref{fig_trienagle}.
The normalized distance $\hat{D'}(\vec{E}q_i,\vec{E}r_j)$ is calculated as the original distance between embeddings $\| \vec{E}q_i - \vec{E}r_j \|_2$ extended with a value of $\beta \sqrt{ \hat{C}(\vec{E}q_i) }$ in a perpendicular direction to the vector $\vec{E}q_i - \vec{E}r_j$.
During the competition, we discovered that we can use this fact to normalize the query embedding $\vec{E}q_i$ by moving them way ($\beta \sqrt{ \hat{C}(\vec{E}q_i) }$ units) from the reference samples to produce new normalized query embeddings $\vec{\hat{E}}q_i$, as it is shown in Figure \ref{fig_trienagle}. 

Since the competition scores samples using Euclidean distances and our ensemble outputs are L2 normalized, by re-scaling the embedding we have one additional degree of freedom that could be used to normalize the query embeddings.
We proposed two query normalization methods, which are described in the following sections.

\subsubsection{Method 1: Escaping from the sphere}
This method increases the length of the query vector we want to normalize, while maintaining its direction in the embedding space.
We select the three training embeddings closest to the query embedding in order to calculate the normalization distance $\hat{C}(\vec{E}q_i)$ using equation \ref{eq_norm6}. After that, we re-scale $\vec{E}q_i$.

\begin{figure}[t!]
	\begin{center}
		\includegraphics[width=1.0\columnwidth]{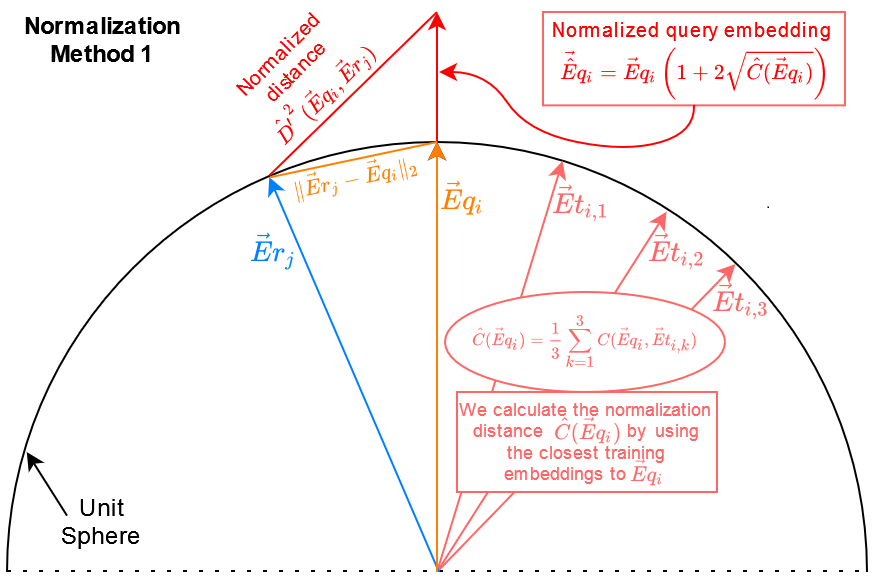}
	\end{center}
	\vspace*{-5mm}
	\caption{\small Geometrical interpretation of the normalization method 1.}
	\label{fig_norm1}
\end{figure}

As it is shown in Figure \ref{fig_norm1}, this procedure guarantees that by moving the query embedding outside the unit sphere, it is also moving away from all the reference embeddings as required by the triangular relationship \ref{eq_norm5}.

Below, we show the equations used to calculate the normalized query embedding $\hat{C}(\vec{E}q_i)$:

\begin{equation}
\hat{C}(\vec{E}q_i)  = \frac{1}{3}\sum_{k=1}^{3} C(\vec{E}q_i,\vec{E}t_{i,k})
\end{equation}

\begin{equation}
\vec{\hat{E}}q_i = \vec{E}q_i \left( 1 + \beta \sqrt{ \hat{C}(\vec{E}q_i) } \right)
\end{equation}

Where:
\begin{itemize}
\item $\vec{E}q_i$: is the embedding of the \emph{i-th} query image.
\item $\vec{E}t_{i,k}$: is the embedding of the \emph{k-th} most similar training image to $\vec{E}q_i$.
\item $C(\vec{E}q_i,\vec{E}t_j)$: is the cosine similarity between $\vec{E}q_i$ and $\vec{E}t_j$.
\item $\hat{C}(\vec{E}q_i)$: is the similarity used to normalize $\vec{E}q_i$.
\item $\beta$: is a constant factor (we used $\beta=2.0$).
\item $\vec{\hat{E}}q_i$: is the normalized embedding of the \emph{i-th} query image.
\end{itemize}

\subsubsection{Method 2: Using training images}

\begin{figure}[h!]
	\begin{center}
		\includegraphics[width=1.0\columnwidth]{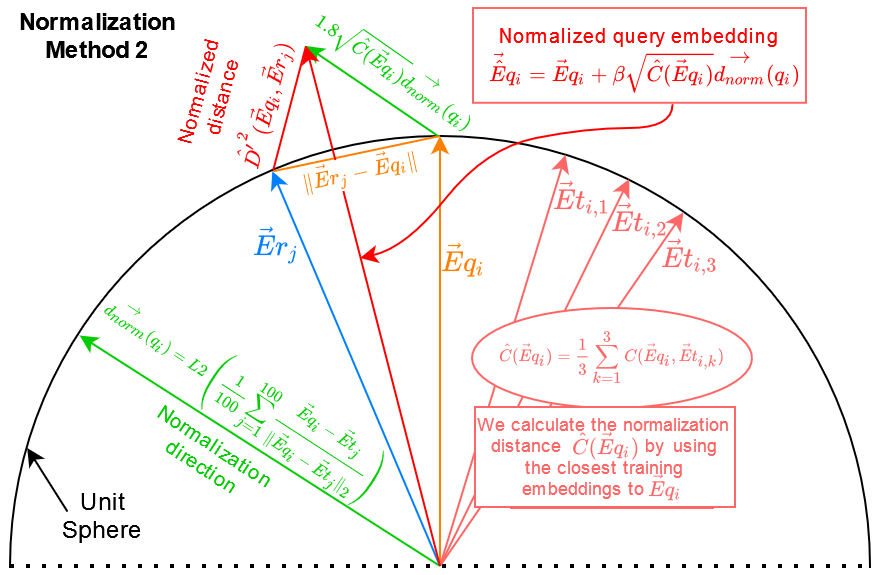}
	\end{center}
	\vspace*{-5mm}
	\caption{\small Geometrical interpretation of the normalization method 2.}
	\label{fig_norm2}
\end{figure}

In Method 1, we proposed $\vec{\hat{E}}q_i$ to be collinear to $\vec{E}q_i$.
However, we can still find a better direction to move $\vec{E}q_i$ using some training samples.
This method proposes to find such direction employing the 100 closest training embeddings to $\vec{E}q_i$.

The idea behind this method is to move $\vec{E}q_i$ away from the 100 most similar training embeddings.
For this normalization method to work, we assume that the distribution of the closest training embeddings is similar to the distribution of the closest reference embeddings.

Below, we show the equations used to calculate the normalized query embedding $\hat{C}(\vec{E}q_i)$:

\begin{equation}
\hat{C}(\vec{E}q_i)  = \frac{1}{3}\sum_{k=1}^{3} C(\vec{E}q_i,\vec{E}t_{i,k})
\end{equation}

\begin{equation}
\vec{d}(q_i) = \frac{1}{100}\sum_{k=1}^{100} \frac{\vec{E}q_i - \vec{E}t_{i,k}}{ \| \vec{E}q_i - \vec{E}t_{i,k}  \|_2}
\end{equation}

\begin{equation}
\vec{d_{norm}}(q_i) = \frac{\vec{d}(q_i)}{\|\vec{d}(q_i)\|_2}
\end{equation}

\begin{equation}
\vec{\hat{E}}q_i = \vec{E}q_i + \beta \sqrt{ \hat{C}(\vec{E}q_i) } \vec{d_{norm}}(q_i)
\end{equation}

Where:
\begin{itemize}
\item $\vec{E}q_i$: is the embedding of the \emph{i-th} query image.
\item $\vec{E}t_{i,k}$: is the embedding of the \emph{k-th} most similar training image to $\vec{E}q_i$.
\item $C(\vec{E}q_i,\vec{E}t_j)$: is the cosine similarity between $\vec{E}q_i$ and $\vec{E}t_j$.
\item $\vec{d}(q_i)$: is the mean training-query direction over all 100 closest training embedding to $\vec{E}q_i$.
\item $\vec{d_{norm}}(q_i)$: is the L2-normalized version of $\vec{d}(q_i)$.
\item $\hat{C}(\vec{E}q_i)$: is the similarity used to normalize $\vec{E}q_i$.
\item $\beta$: is a constant factor (we used $\beta=1.8$).
\item $\vec{\hat{E}}q_i$: is the normalized embedding of the \emph{i-th} query image.
\end{itemize}

Figure \ref{fig_norm2} shows that this time $\vec{\hat{E}}q_i$ is not col-linear to $\vec{E}q_i$. $\vec{E}q_i$ is moving away from the 100 closest training embeddings, by assuming that the distributions of the closest reference and training embeddings are similar, $\vec{E}q_i$ is also moving away from the closest reference embeddings as required by relationship \ref{eq_norm5}.
 
Method 2 proved to be the most effective way to normalize the query embeddings. At Phase 2 of the competition, by using this method we could improve our $\mu AP$ score from $0.53$ to $0.59$.

\section{Conclusion}

In this paper, we presented our solution for The Facebook 2021 Image Similarity Challenge. 
Our solution uses different CNN backbones to produce the embeddings for each of the input images. 
The backbones were trained using an ArcFace head. 
We presented a way to ensemble the different models and two new methods to perform query embeddings normalization.
Moreover, we presented a way to perform rapid training for our models by using a progressive training scheme.
After ensembling our models, we reached a final score of $\mu AP =0.59$ which leaded us to win the 2nd place in Phase 2 of the competition.

%\newpage
\bibliographystyle{abbrv}
\bibliography{refs}
\end{document}